\documentclass[11pt,a4paper]{article}
\usepackage{authblk}

\usepackage[hyperref]{acl2021}
\usepackage{times}
\usepackage{latexsym}

\usepackage{microtype}

\aclfinalcopy 

\usepackage{multirow}
\usepackage{amssymb}
\usepackage{mathtools}
\usepackage{subcaption}
\usepackage{algorithm}
\usepackage{algpseudocode}

\title{OutFlip: Generating Out-of-Domain Samples for Unknown Intent Detection with Natural Language Attack}

\author[1,2]{ DongHyun Choi}
\author[1]{ Myeong Cheol Shin}
\author[1]{ EungGyun Kim}
\author[2]{ Dong Ryeol Shin}

\affil[1]{ Kakao Enterprise, Pangyo, South Korea}
\affil[ ]{\textit {\{heuristic.c,index.sh.jason.ng\}@kakaoenterprise.com}}
\affil[2]{Sungkyunkwan University, Suwon, South Korea}
\affil[ ]{\textit {drshin@skku.edu}}

\date{}

\begin{document}
\maketitle
\begin{abstract}
Out-of-domain (OOD) input detection is vital in a task-oriented dialogue system since the acceptance of unsupported inputs could lead to an incorrect response of the system. This paper proposes OutFlip, a method to generate out-of-domain samples using only in-domain training dataset automatically. A white-box natural language attack method HotFlip is revised to generate out-of-domain samples instead of adversarial examples. Our evaluation results showed that integrating OutFlip-generated out-of-domain samples into the training dataset could significantly improve an intent classification model's out-of-domain detection performance\footnote{Soon will be available at https://github.com/kakaoenterprise/OutFlip}.
  
 \end{abstract}

\section{Introduction}
Intent classification is crucial for task-oriented dialogue systems such as Google DialogFlow or Amazon Lex. It is vital for an intent classifier not only to map an input utterance into the correct label but also to detect \textit{out-of-domain} (OOD) inputs. An accepted OOD input will lead the dialogue system to give erroneous responses.

Approaches for OOD detection in text classification could be classified into two major categories. Outlier detection approaches \citep{Fei:16,hendrycks:17,doc-Shu:17,LMCL-Lin:19,gaussian-Yan:20,Xu:2020} try to find out the boundaries of known classes in feature space. They need no labeled OOD dataset, but it is hard for them to deal with boundary cases.  $(n+1)$-way classification approaches \citep{kim:18,larson:19,ryu:18,zheng:20} train classifiers for OOD detection using (pseudo-)labeled OOD samples. In practice, it is difficult and expensive to collect a large number of labeled OOD samples with an open-world environment.

This paper proposes OutFlip, a method to generate OOD samples from in-domain training dataset automatically. For a given training dataset $T$ and a reference intent classification model $M$ which is trained with $T$, the OutFlip generates a set of OOD samples $O$. The generated OOD samples $O$ could be used to train $M$ iteratively to improve its OOD detection performance. Since the OutFlip does not require any modifications to the model architecture, it could be used with other OOD detection approaches to further improve the OOD detection performance.

 The generated OOD samples should satisfy two conditions. First, they should be ``hard-enough''; if the generated examples are too easy to distinguish from in-domain intents, they will be useless in training the OOD detector. Second, they should not belong to any in-domain intents. With a given reference model $M$ and a set of in-domain labels $I$, this could be considered as finding a sentence $\mathbf{x}_o$ with truth label $y \not \in I$ and model classification $y' \in I$. In this point of view, the OOD sample generation task could be considered as a variant of natural language attack on model $M$; the goal of natural language attack on $M$ is to find a $\mathbf{x}_a$ with truth label $y \in I$ and model classification $y' \ne y$. We revised HotFlip \citep{hotflip}, a natural language attack method, to generate such OOD samples.
  
Our evaluation results showed that the generated OOD samples could significantly improve the OOD detection performances of the reference models. We also showed that applying OutFlip with other OOD detection approaches could further improve the model's OOD detection performance. The evaluation results also suggest that the generated OOD samples could train the models other than the reference model to improve their OOD detection performances.

Our contributions are summarized as follows:

\begin{itemize}
\item We proposed OutFlip, a simple and efficient OOD sample generation method using only in-domain training samples.
\item We experimentally showed the effectiveness of our proposed approach using the intent classification benchmarks.
\item We showed that the generated OOD samples could also improve the OOD detection performances of models other than the reference model.
\end{itemize}

\section{Related Work}
\label{related}

Previous OOD detection works could be classified into two major categories. Outlier detection approaches find boundaries of known classes in feature space. \citet{Fei:16} computes a center for each class and transforms each document into a vector of similarities to the center. A binary classifier is built using the transformed data for each class. For deep learning-based systems, \citet{hendrycks:17} proposed the baseline of using softmax score as a threshold. \citet{doc-Shu:17} trained the intent classifier using the sigmoid function and used the standard distribution to set each class's score threshold. \citet{LMCL-Lin:19} first trained the classifier using Large Margin Cosine Loss (LMCL) \citep{lmcl-orig}, and applied Local Outlier Factor \citep{lof} to detect the OOD inputs. \citet{gaussian-Yan:20} proposed a semantic-enhanced Gaussian mixture model to gather vectors of the same classes closely. \citet{Xu:2020} calculated the mean and covariance of training samples for each class and used Mahalanobis distance as a distance function. 

 $(n+1)$-way classification approaches train the intent classifier with one additional class, where $(n+1)$-th class represents the unseen intent. \citet{kim:18} proposed joint learning for in-domain and out-of-domain speeches. \citet{larson:19} manually collected OOD samples to train intent classifiers. \citet{ryu:18} generated OOD feature vectors using generative adversarial network \citep{gan} to train an OOD detector. Since the approach proposed in \citet{ryu:18} only works on continuous feature space, it highly depends on the feature encoder, which transforms inputs into feature vectors. \citet{zheng:20} also generated OOD feature vectors, but they also used unlabeled examples to enhance classification performance further. Although the $(n+1)$-way classification approaches are easy to adopt without modification in the classification model, it is incredibly costly and time-consuming to collect the appropriate OOD samples. 
 
 The proposed OutFlip automatically generates OOD samples using the only in-domain training set, significantly reducing the cost of manually collecting OOD samples. Also, the OutFlip does not depend on the feature encoder.
   
 The goal of adversarial attack in text classification is to fool a given text classification model $M$, by generating an adversarial example $\textbf{x}_a$ with truth label $y$ and model classification $y' \ne y$.  Many successful attacks first take a correctly classified example $\textbf{x}$ and replace its important words or characters to get an adversarial sample $\textbf{x}_a$. In a white-box scenario, the attacker has access to the target model's structure; thus, the important word or characters could be easily selected by inspecting the gradient of model $M$. HotFlip \citep{hotflip} estimates the best change of characters by maximizing the first-order approximation of the change in the loss. 
 
 In a black-box scenario, the attacker is not aware of the model or training data; the attacker is only capable of querying the target model with supplied inputs and obtaining the output predictions and their confidence scores. \citet{alzantot:18} randomly selects a word from sentence $\textbf{x}$ and selects a suitable replacement word that has a similar semantic meaning. \citet{jin:20} proposed a word importance score, which is used to find the word to be replaced. \citet{bert-attack} applied BERT \citep{devlin2018bert} pre-trained language model to find a replacement word.
   
 The proposed OutFlip first extracts important words using the algorithm proposed in \citet{jin:20}, and applies a variant of HotFlip \citep{hotflip} to generate OOD samples which are hard to distinguish from the in-domain intents by the given reference model.
 
\section{Proposed Approach}
In this section, the proposed OOD sample generation approach OutFlip is described in more detail.
\subsection{HotFlip}
We first introduce the white-box adversarial example generation method HotFlip \citep{hotflip}. Let $M$ be a text classification model, $V$ be the word vocabulary set, $\mathbf{x} = \{x_1; ...; x_n\}$ be a sentence with $n$ words where $x_i \in \{0,1\}^{|V|}$ denotes one-hot vector representing the $i$-th word, and $L_M(\mathbf{x}, \mathbf{y})$ be the loss of $M$ on input \textbf{x} with true output $\mathbf{y}$. For a given sentence $\mathbf{x}$, a flip of the $i$-th word from $w_a$ to $w_b$ is represented by the following vector:

\begin{equation}
\begin{aligned}
\vec{v}_{ib} = (\vec{0}; ..;( 0,..,-1,..,1,0,.., 0)_{i};..;\vec{0} )
\end{aligned}
\end{equation}

where -1 and 1 are in the corresponding positions for words $w_a$ and $w_b$ in the word vocabulary, respectively. A first-order approximation of the change in loss $L_M(\mathbf{x},\mathbf{y})$ can be obtained from a directional derivative along this vector:

\begin{equation}
\begin{aligned}
\nabla_{\vec{v}_{ib}}L_M(\mathbf{x},\mathbf{y}) = \nabla_{x} L_M(\mathbf{x},\mathbf{y})^T \cdot \vec{v}_{ib}  
\end{aligned}
\end{equation}

Then, the HotFlip chooses the vector with the biggest increase in loss:

\begin{equation}
\begin{aligned}
\max \nabla_{x}L_M(\mathbf{x},\textbf{y})^T& \cdot \vec{v}_{ib} = \max_{ib}\frac{ \partial L_M^{(b)}}{\partial x_i} - \frac{ \partial L_M^{(a)}}{\partial x_i}& \\
\text{subject to } &sim(w_a,w_b) \geq T_{sim} \\
\text{and } &POS(w_a) = POS(w_b)
\end{aligned}
\label{eq:hotflip}
\end{equation}

where $T_{sim}$ is a similarity threshold between two words, and $POS(w_a)$ is the Part-of-Speech tagging of $w_a$. The two constraints are added to ensure that $\mathbf{x}_a$ is semantically similar to the original input $\mathbf{x}$. With equation \ref{eq:hotflip}, the HotFlip could determine the flip position $i$ and the replacement word $w_b$.

\subsection{OutFlip}

\begin{algorithm}[h]
\textbf{Input} In-domain training corpus $T$ = \{$\mathbf{x}_1$, ..., $\mathbf{x}_t$\}, \\
\hspace*{27pt} in-domain labels $Y$, vocabulary $V$, reference \\
\hspace*{27pt} model $M$, similarity threshold $T_{sim}$ \\
\textbf{Output} A set of OOD samples $O$ 
\begin{algorithmic}[1]
\For{ $ y \in Y$ }
\For { $\mathbf{x}$ with truth label $y$}
\State Calculate $w_M(\mathbf{x})$ 
\EndFor
\State $C_T(y) \gets$ Top 5 most frequent $w_M(\mathbf{x})$
\EndFor
\State
\For {$\mathbf{x} \in T$} 
\State $y \gets$ truth label of $\mathbf{x}$
\If {$w_M(\mathbf{x}) \in C_T(y)$}
\State $i \gets$ the position of $w_M(\mathbf{x})$
\State Sort $V$ in ascending order of  $\frac{ \partial L_M^{(b)}}{\partial x_i}$ 
\State Candidate $\gets$ Top 1\% of $V$
\State Remain only words whose similarity \hspace*{40pt} with $w_M(\mathbf{x})$ is less than $T_{sim}$
\State Randomly select $w_b$ among candidates
\State Replace $w_M(\mathbf{x})$ with $w_b$ to get $\mathbf{x}_o$
\If { $M$ classifies $\mathbf{x}_o$ to $y$}
\State Add $\mathbf{x}_o$ to $O$
\EndIf
\EndIf
\EndFor
\end{algorithmic}
 \caption{OutFlip}
 \label{alg:OutFlip}
\end{algorithm}

For a given reference model $M$ and an in-domain sample $\mathbf{x}$ with true output $\mathbf{y}$, the main idea of OutFlip is to flip the most important word of $\mathbf{x}$, $w_M(\mathbf{x})$, to a semantically different word, while \textit{minimizing} the change of loss $L_M(\mathbf{x}, \mathbf{y})$. By doing so, the OutFlip expects to get a sample $\mathbf{x}_o$ whose truth label is different from the truth label of $\mathbf{x}$, while the model classifications of $\mathbf{x}_o$ and $\mathbf{x}$ are the same.

The word importance score proposed in \citet{jin:20} is defined as follows:

\begin{equation}
\begin{aligned}
I_{x_i}(M, \mathbf{x}) = o_y(M, \mathbf{x}) - o_y(M, \mathbf{x}_{\backslash x_i})
\end{aligned}
\end{equation}

where $y$ is the truth label of $\mathbf{x}$, $o_y(M, \mathbf{x})$ is the logit output of the target model $M$ for label $y$, and $\mathbf{x}_{\backslash x_i}$ is the sentence after masking $x_i$. The most important word $w_M(\mathbf{x})$ is defined as the word with the largest importance score in $\mathbf{x}$.

For each in-domain label $y$ of the training dataset $T$, we define Core Class Token (CCT) $C_T(y)$ as the top 5 most frequent $w_M(\mathbf{x})$ among the training samples with truth label $y$. Since the importance score is calculated based on the reference model $M$, the OutFlip could select a wrong token as the most important token due to the model error. If the OutFlip flips such a word, the generated sentence's truth label will remain unchanged, leading to an erroneous OOD sample. To prevent such case, the OutFlip simply disregards $\mathbf{x}$ during OOD generation process if $w_M(\mathbf{x}) \not \in C_T(y)$. 

In summary, the OutFlip chooses the replacement word $w_b$ using the following equation:

\begin{equation}
\begin{aligned}
\min \nabla_{x}L_M(\mathbf{x}, \mathbf{y})^T & \cdot \vec{v}_{ib} = \min_{ib}\frac{ \partial L_M^{(b)}}{\partial x_i} - \frac{ \partial L_M^{(a)}}{\partial x_i}& \\
\text{subject to } &sim(w_a,w_b) \leq  T_{sim} \\
\text{and } &w_a = w_M(\mathbf{x} )\\
\text{and }& w_a \in C_T(y)  
\end{aligned}
\end{equation}

Since we do not need the generated OOD samples to be fluent, the part-of-speech condition is removed. The OutFlip randomly chooses $w_b$ among the top 1\% of the vocabulary in the ascending order of the loss change to generate more diverse samples. We used cosine similarity as the similarity measure.

The truth label of the generated sample $\mathbf{x}_o$ could be an in-domain label different from $y$ by chance. The OutFlip checks the model classification result of $\mathbf{x}_o$ to see if it remains the same as $\mathbf{x}$. If the classification result changes, the OutFlip disregards $\mathbf{x}_o$. Algorithm \ref{alg:OutFlip} shows the pseudocode of the proposed OutFlip.

\subsection{Iteratively Populating OOD samples}
\begin{figure}
\centering
\includegraphics[width=0.5\textwidth]{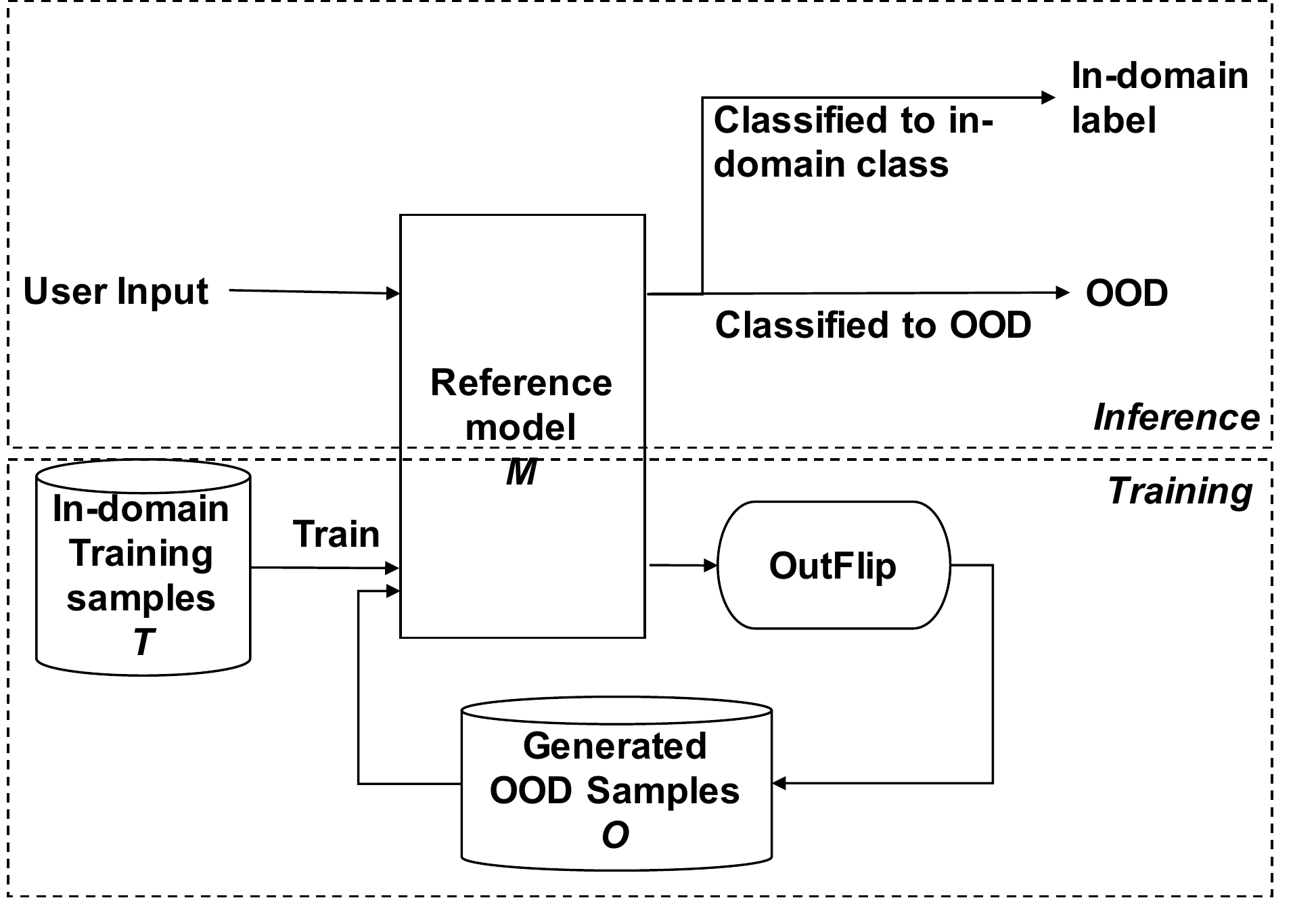}
\caption{Applying the OutFlip to iteratively train the reference model $M$ with newly generated OOD samples. }
\label{fig:architecture}
\end{figure}

The reference model $M$ could be iteratively trained with the generated OOD samples to improve its OOD detection performance. Figure \ref{fig:architecture} shows the overall framework. For each iteration, the set of generated OOD samples $O$ is randomly split into training and dev set and used for the next train iteration.

Since the OutFlip does not require any change in model architecture, the OutFlip could be applied independently with other OOD detection algorithms that require modifications on model architecture or loss function, such as \citet{doc-Shu:17} or \citet{LMCL-Lin:19}.  In such cases, those OOD detection algorithms are applied to the examples classified as in-domain to filter out the OOD samples further.  

\section{Experiments}
In this section, experimental settings and evaluation results are shown.

\subsection{Datasets}
\begin{table}
\centering
\begin{tabular}{lccc} \hline
Dataset&\textbf{ATIS}&\textbf{SNIPS}&\textbf{Kakao} \\ \hline
Language&English&English&Korean \\
Vocab Size & 938& 12,054 &22,831 \\ 
Avg. Length & 11.21 & 9.36 & 8.88 \\ 
\# Train & 4,478 & 13,784 & 90,692 \\ 
\# Dev & 500 & 700 & 11,310 \\ 
\# Test & 893 & - & 12,711 \\ 
\# of Classes & 18 & 7 & 48 \\ 
Is Balanced& X & O & O \\ \hline
\end{tabular}
\caption{Dataset statistics.}
\label{tbl:dataset}
\end{table}

Experiments are conducted on 3 real task-oriented dialogue datasets, SNIPS \cite{snips}, ATIS \cite{atis} and Kakao dialogue corpus\footnote{Since the Kakao dataset is not publicly available, we contacted the authors to get the Kakao dataset and Korean GloVe vectors.} \cite{kakao}. SNIPS and ATIS are well-known English benchmarks. Kakao dialogue corpus is a Korean intent classification benchmark. We evaluated the proposed OutFlip with the Kakao dataset to see if it could be applied to different languages. Table \ref{tbl:dataset} summarizes the statistics of the datasets.  The ATIS dataset is highly imbalanced; more than 70\% samples belong to one class, while three classes have less than 10 samples. The SNIPS and Kakao datasets are relatively balanced.




Since the SNIPS dataset does not have a test set, we randomly selected 30\% of the training set and used them as the test set.

\subsection{Baselines}
\begin{table*}[t]
\centering
\begin{tabular}{l|ccc|ccc|ccc} \hline
Dataset&\multicolumn{3}{|c|}{\textbf{ATIS}} & \multicolumn{3}{c|}{\textbf{SNIPS}} & \multicolumn{3}{c}{\textbf{Kakao}} \\ 
\% of known intents&25\%&50\%&75\%&25\%&50\%&75\%&25\%&50\%&75\% \\ \hline
\textbf{MSP$_{cnn}$}& 64.61&71.07&62.00&34.24&63.26&73.14&66.23&82.32& 89.60 \\
\textbf{MSP$_{lstm}$}& 65.82&71.52&61.18&34.12&63.30&73.28&64.23&81.37& 88.28 \\
\textbf{DOC$_{cnn}$} &61.99&57.82& 38.46 & 49.66&70.76&77.16&71.61&84.72& \textbf{90.55} \\
\textbf{DOC$_{lstm}$} & 62.76& 58.15&38.35& 49.77&71.11&77.37&63.36&78.86& 85.22\\
\textbf{LMCL$_{cnn}$}  &71.67&74.89&68.73&61.51&\textbf{84.37}&88.31&80.27&\textbf{87.15}&\textbf{90.60} \\
\textbf{LMCL$_{lstm}$} & 72.25&\textbf{77.90}&\textbf{73.18}&69.52&83.32&87.53&76.31&\textbf{85.79}&89.39 \\  \hline

\textbf{OutFlip$_{cnn}$} & \textbf{74.18} &\textbf{79.23} & \textbf{69.37} & \textbf{79.20} & \textbf{84.25} & \textbf{88.99} & \textbf{81.78} & 85.45 & 86.85     \\
\textbf{OutFlip$_{lstm}$} & \textbf{73.85} & 74.50 & 68.30 & \textbf{79.26} & 84.00 & \textbf{88.67} & \textbf{81.96}& 84.64& 86.42 \\ \hline
\end{tabular}
\caption{Comparisons of the OutFlip and previous OOD detection works. The top 2 results for each metric are marked in bold. $T_{sim}$ is set to 0.3, and the OutFlip is applied for three iterations.}
\label{tbl:result}
\end{table*}

We implemented two sentence encoders to show the generality of the proposed approach. LSTM \citep{lstm}-based encoder applies one-layer BiLSTM with output dimension 128 on the word embeddings of the given input; a self-attention layer with attention dimension 10 is followed to get the feature vector. CNN-based model applies the algorithm proposed in \citet{cnn-sen}. More precisely, one-dimensional convolutions with kernel sizes 2, 3, 4, 5 and filter size 32 are applied on top of the word embeddings. The results are max-pooled to get the feature vector. For both encoders, a dense layer is applied to the feature vector to get the logit of each class.

We also implemented three baseline OOD detection systems, as follows:

\begin{enumerate}
\item \textbf{Maximum Softmax Probability (MSP)} \citep{hendrycks:17} considers the maximum softmax probability of a sample as the rejection score. If the probability is below a certain threshold, the sample is classified as OOD. We used the threshold of 0.5, as the authors suggested.
\item \textbf{Deep Open Classification (DOC)} \citep{doc-Shu:17} replaces softmax with sigmoid activation as the final layer to calculate the score for each class separately. It also calculates the threshold for each class through a statistical approach.
\item \textbf{Large Margin Cosine Loss (LMCL)} \citep{LMCL-Lin:19} replaces the softmax loss with large margin cosine loss \citep{lmcl-orig}, to force the model to maximize inter-class variance and minimize intra-class variance. After training, it applies Local Outlier Factor (LOF) \citep{lof} on training features vectors to detect outliers as OOD. We set the scaling factor $s=30$ and cosign margin $m = 0.35$, following the authors.
\end{enumerate}

By combining two feature encoders and three baseline OOD detection systems, we implemented eight baseline reference models, six with an OOD detection system and two without.

\subsection{Experimental Setup}

\begin{table}
\centering
\begin{tabular}{lc} \hline
Pre-trained Embeddings& Accuracy \\ \hline
\textbf{GloVe} \citep{glove} & 83.17 \% \\
\textbf{Korean GloVe} \citep{kakao} & 51.33 \% \\ \hline
\end{tabular}
\caption{Evaluation results of GloVe embeddings on the language-independent set of word analogy corpus.}
\label{tbl:wordanalogy}
\end{table}

\begin{table*}[t]
\centering
\begin{tabular}{l|ccc|ccc|ccc} \hline
Dataset&\multicolumn{3}{|c|}{\textbf{ATIS}} & \multicolumn{3}{c|}{\textbf{SNIPS}} & \multicolumn{3}{c}{\textbf{Kakao}} \\ 
\% of known intents&25\%&50\%&75\%&25\%&50\%&75\%&25\%&50\%&75\% \\ \hline
\multirow{2}{*}{\textbf{MSP$_{cnn}$}}   & 77.04& 74.66 & 62.44 & 78.02 & 84.48& 89.66& 82.65 & 85.74 & 87.26 \\
&$^{\textbf{(+12.43)}}$& $^{\textbf{(+3.59)}}$ & $^{\textbf{(+0.44)}}$ &$^{\textbf{(+43.78)}}$  & $^{\textbf{(+21.22)}}$ & $^{\textbf{(+16.52)}}$ & $^{\textbf{(+16.42)}}$ &$^{\textbf{(+3.42)}}$  & $^{\underline{\textbf{(-2.34)}}}$ \\

\multirow{2}{*}{\textbf{MSP$_{lstm}$}} & 71.37 & 76.17 & 61.97 & 79.88 & 83.98 & 88.92 & 82.22 & 85.21 & 86.25 \\
&$^{\textbf{(+5.55)}}$& $^{\textbf{(+4.65)}}$& $^{\textbf{(+0.79)}}$& $^{\textbf{(+45.76)}}$& $^{\textbf{(+20.68)}}$& $^{\textbf{(+15.64)}}$& $^{\textbf{(+17.99)}}$&$^{\textbf{(+3.84)}}$ & $^{\underline{\textbf{(-2.03)}}}$ \\

\multirow{2}{*}{\textbf{DOC$_{cnn}$}}   & 73.45 & 59.03 & 41.24 & 80.72 & 85.05 & 89.42 & 82.20 & 85.45 & 86.44 \\
&$^{\textbf{(+11.46)}}$& $^{\textbf{(+1.21)}}$&$^{\textbf{(+2.78)}}$&$^{\textbf{(+31.06)}}$ &$^{\textbf{(+14.29)}}$ &$^{\textbf{(+12.26)}}$ &$^{\textbf{(+10.59)}}$ &$^{\textbf{(+0.73)}}$ & $^{\underline{\textbf{(-4.11)}}}$ \\

\multirow{2}{*}{\textbf{DOC$_{lstm}$}}  & 70.20 & 59.69 & 41.09 & 81.99 & 85.08 & 89.69 & 80.14 & 81.28 & 80.78 \\
&$^{\textbf{(+7.44)}}$& $^{\textbf{(+1.54)}}$ & $^{\textbf{(+2.74)}}$ & $^{\textbf{(+32.22)}}$ & $^{\textbf{(+13.97)}}$ & $^{\textbf{(+12.32)}}$ & $^{\textbf{(+16.78)}}$ & $^{\textbf{(+2.42)}}$& $^{\underline{\textbf{(-4.44)}}}$  \\

\multirow{2}{*}{\textbf{LMCL$_{cnn}$}}   &74.13 & 76.35 & 68.82 & 81.04 & 85.90 & 90.10 & 81.81 & 84.75 & 85.74 \\
&$^{\textbf{(+2.46)}}$&  $^{\textbf{(+1.46)}}$ & $^{\textbf{(+0.09)}}$ & $^{\textbf{(+19.53)}}$ & $^{\textbf{(+1.53)}}$ & $^{\textbf{(+1.79 )}}$ & $^{\textbf{(+1.54)}}$ & $^{\underline{\textbf{(-2.40)}}}$ & $^{\underline{\textbf{(-4.86)}}}$  \\
\multirow{2}{*}{\textbf{LMCL$_{lstm}$}}  & 75.44 & 76.37 & 71.34 & 80.86 & 84.56 & 88.76 & 82.24 & 84.19 & 85.35 \\
&$^{\textbf{(+3.19)}}$& $^{\underline{\textbf{(-1.53)}}}$ & $^{\underline{\textbf{(-1.84)}}}$ & $^{\textbf{(+11.34)}}$&$^{\textbf{(+1.24)}}$ &$^{\textbf{(+1.23)}}$ &$^{\textbf{(+5.93)}}$ & $^{\underline{\textbf{(-1.60)}}}$& $^{\underline{\textbf{(-4.04)}}}$ \\ \hline
\end{tabular}
\caption{Evaluation results of the baselines with the OutFlip applied. $T_{sim}$ is set to 0.3, and the OutFlip is applied for three iterations. Small numbers below the macro F1 score represents the performance improvement compared to the same baseline without applying the OutFlip.}
\label{tbl:result2}
\end{table*}

Word embeddings are initialized with GloVe \citep{glove} pre-trained word vectors. We downloaded the pre-trained embeddings containing 1.9M words trained on 42B tokens from the author's homepage. For Korean, Korean pre-trained GloVe embedding vectors proposed in \citet{kakao} are used. The dimensions of both pre-trained embeddings are 300.

We removed some classes from the train/dev set during training and integrated them back during testing, following the evaluation settings of \citet{Fei:16, doc-Shu:17, LMCL-Lin:19}. We varied the number of known intents in the training dataset as 25\%, 50\%, and 75\% of the intents, and used all intents for testing. We randomly select known intents by weighted random sampling without replacement in the training set. Note that the samples belonging to the unknown intents are removed during training and validation.

Following \citet{Fei:16, doc-Shu:17, LMCL-Lin:19}, macro F1 score is used to evaluate the models. For each known intent selection, the F1 score for each class is calculated separately. Then the results are macro-averaged across all classes. We reported the average of 10 random known intent selections for each evaluation.

For each OutFlip iteration, 90\% of the generated OOD samples are added to the training set, and the remaining 10\% are added to the dev set. The populated train/dev sets are used for the next train iteration.

Adam optimizer \citep{adam} with an initial learning rate of 0.001 is used to train the model. The training batch size is set to 128. Exponential learning rate decay with a decay rate of 0.8 is applied for every two epochs. On each epoch, the trained classifier is evaluated against the dev set, and the training stops when the dev accuracy is not improved for five consequent epochs. 

\subsection{Evaluation Results}
\begin{figure*}[ht]
\begin{subfigure}{.33\textwidth}
  \centering
  \includegraphics[width=1.0\linewidth]{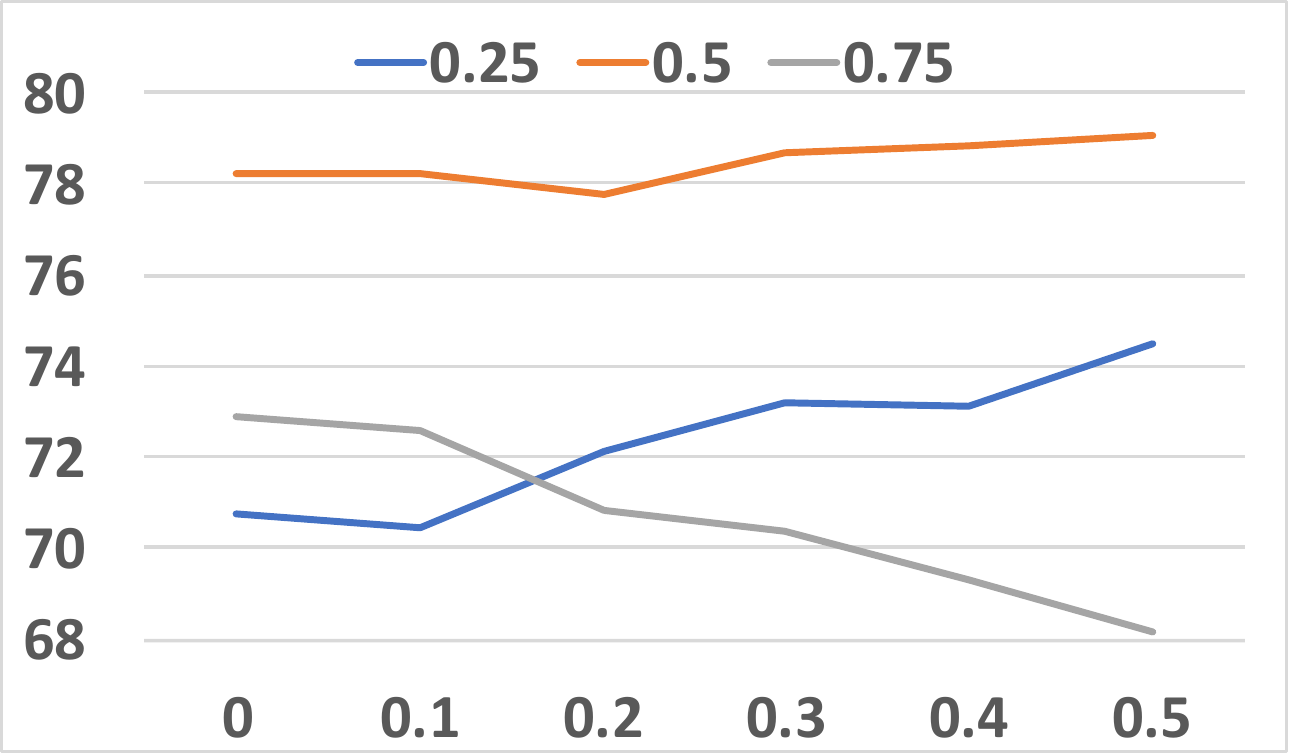}  
  \caption{ATIS dataset}
  \label{fig:sim-first}
\end{subfigure}
\begin{subfigure}{.33\textwidth}
  \centering
  \includegraphics[width=1.0\linewidth]{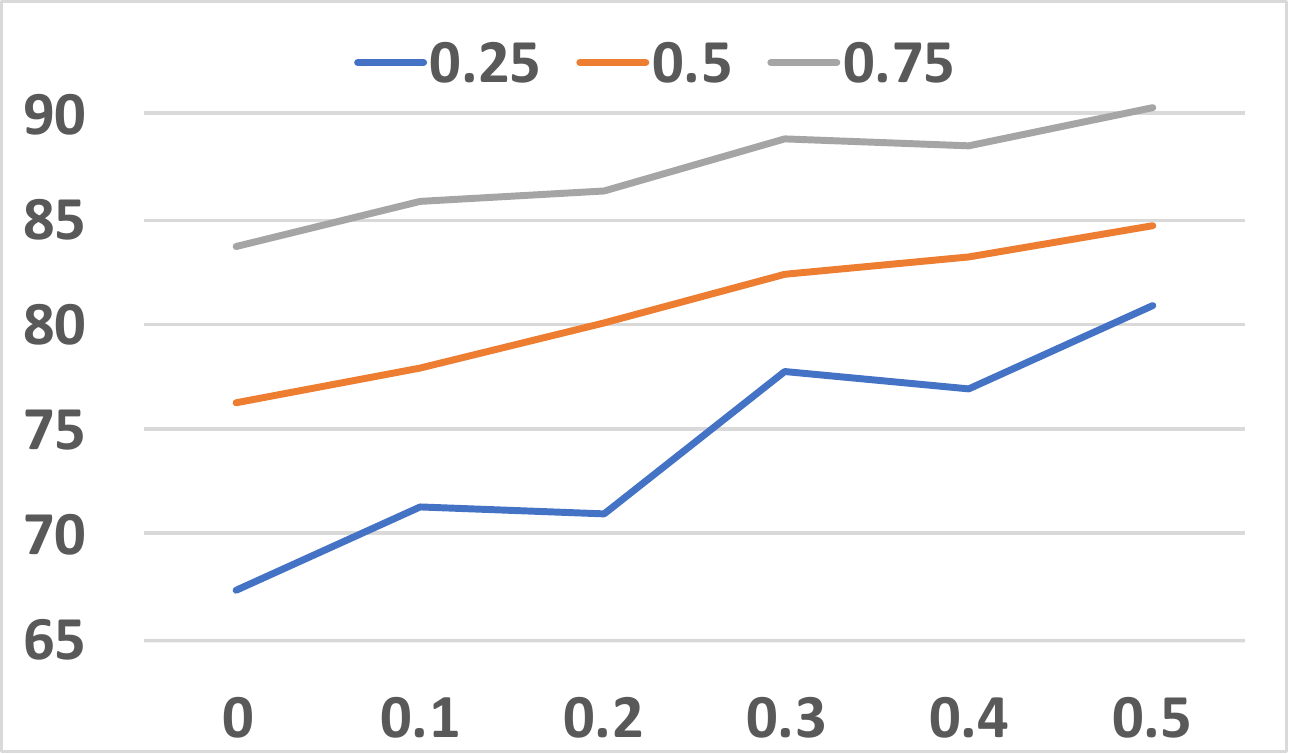}  
  \caption{SNIPS dataset}
  \label{fig:sim-second}
\end{subfigure}
\begin{subfigure}{.33\textwidth}
  \centering
  \includegraphics[width=1.0\linewidth]{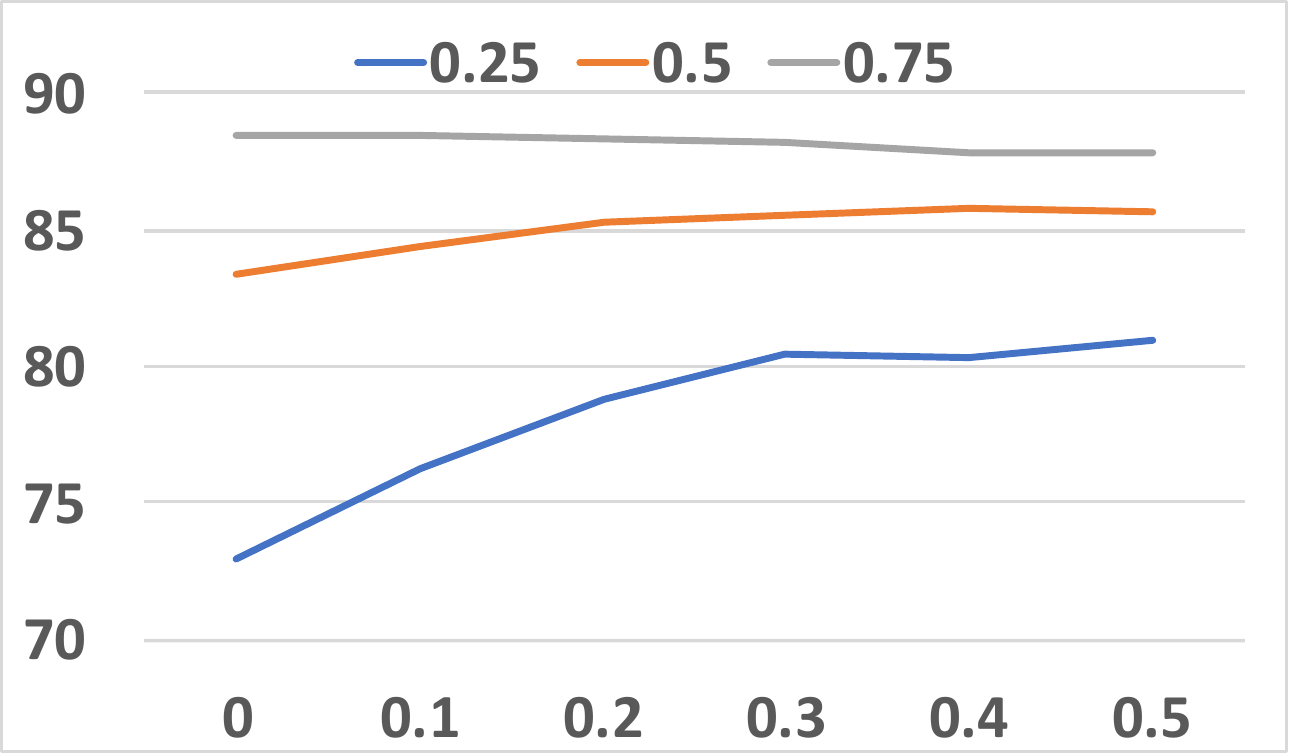}  
  \caption{Kakao dataset}
  \label{fig:sim-third}
\end{subfigure}
\caption{Evaluation results of \textbf{OutFlip$_{cnn}$} with changing $T_{sim}$. We fixed the number of OutFlip iteration to 2.}
\label{fig:test_sim}
\end{figure*}

\begin{figure*}[ht]
\begin{subfigure}{.33\textwidth}
  \centering
  \includegraphics[width=1.0\linewidth]{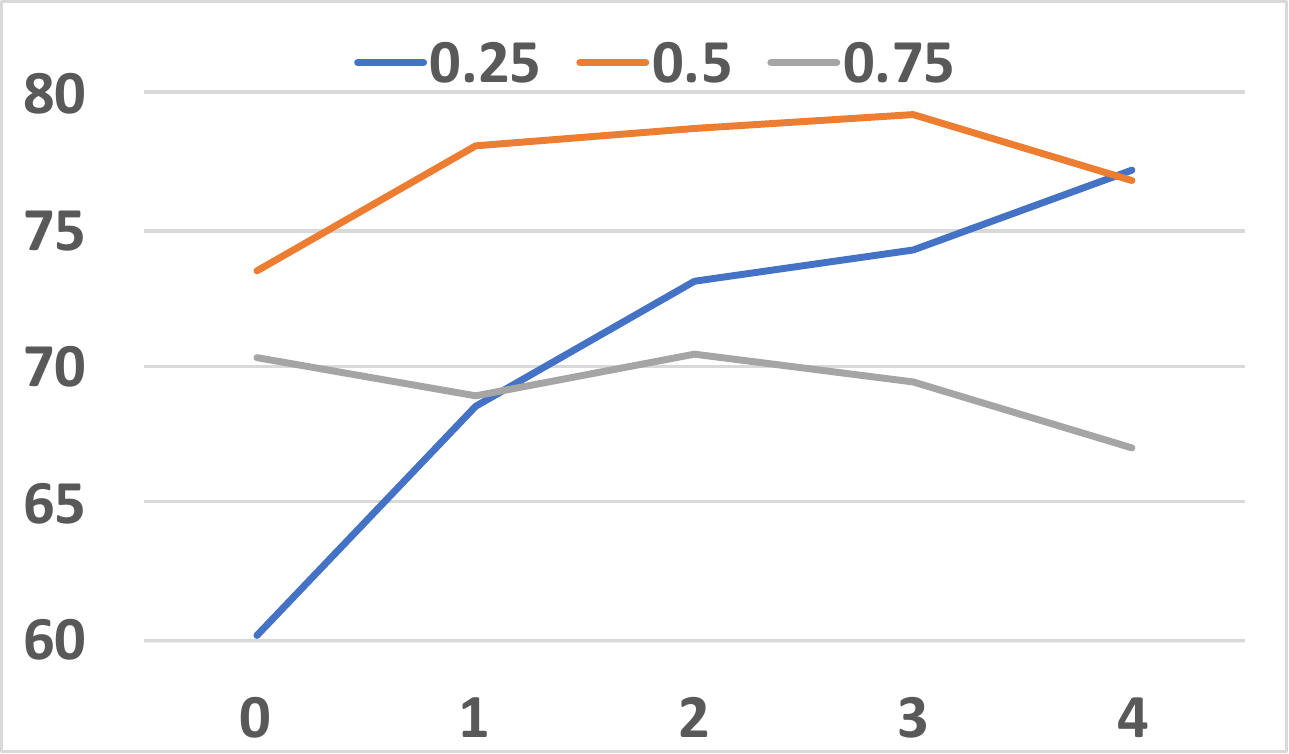}  
  \caption{ATIS dataset}
  \label{fig:sub-first}
\end{subfigure}
\begin{subfigure}{.33\textwidth}
  \centering
  \includegraphics[width=1.0\linewidth]{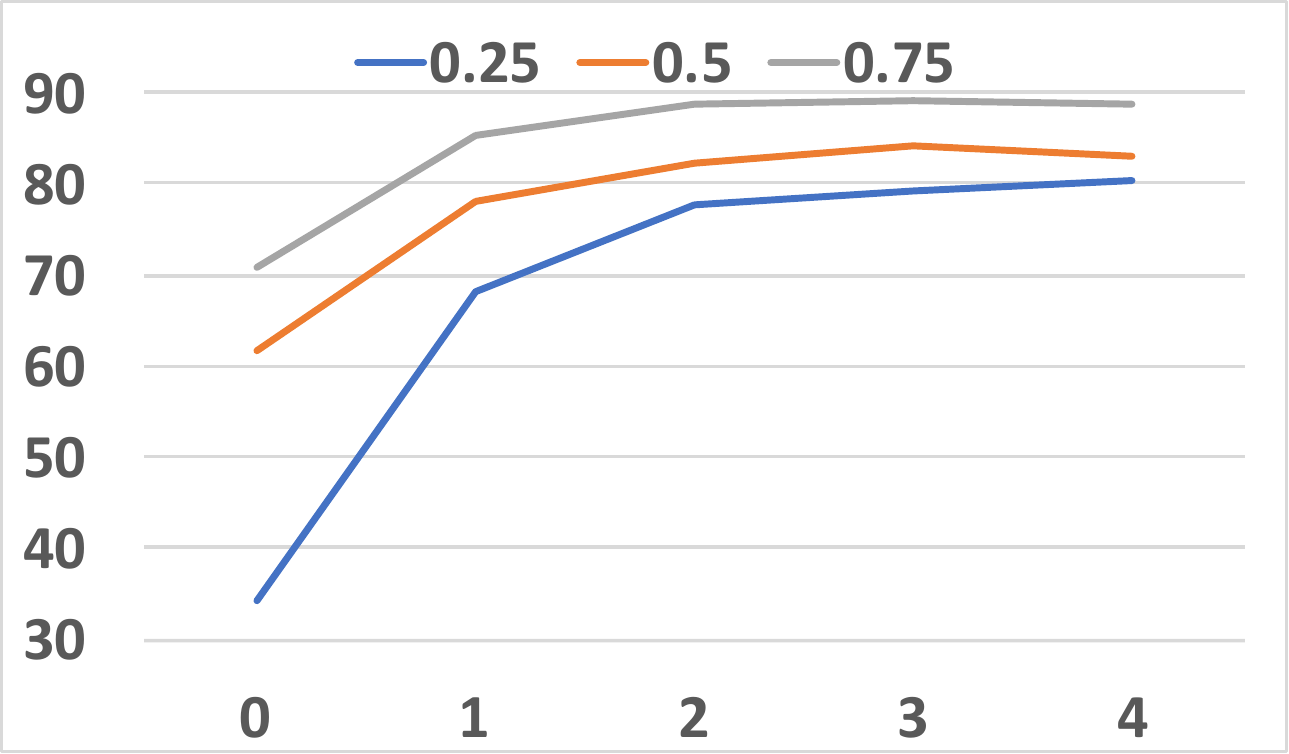}  
  \caption{SNIPS dataset}
  \label{fig:sub-first}
\end{subfigure}
\begin{subfigure}{.33\textwidth}
  \centering
  \includegraphics[width=1.0\linewidth]{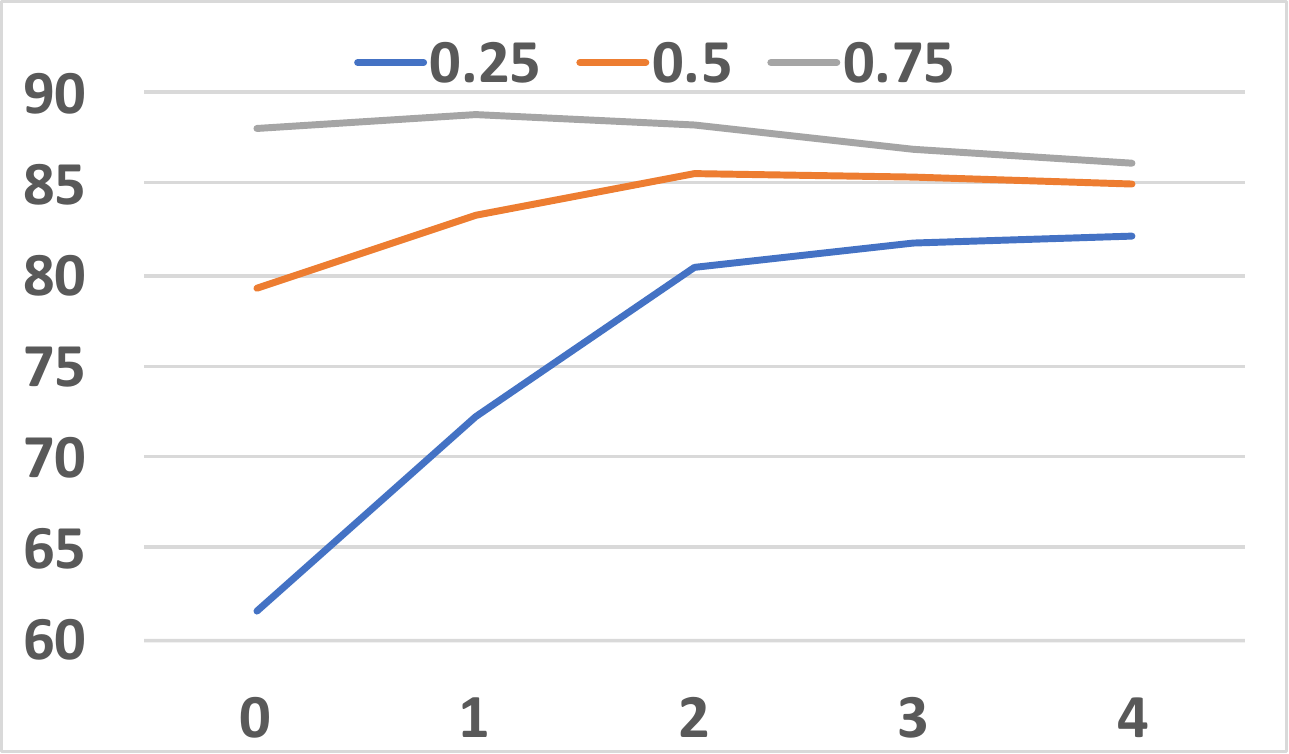}  
  \caption{Kakao dataset}
  \label{fig:sub-first}
\end{subfigure}

\caption{Evaluation results of \textbf{OutFlip$_{cnn}$} with changing OutFlip iteration number. We fixed $T_{sim}$  to 0.3.}
\label{fig:test_attack}
\end{figure*}

Table \ref{tbl:result} shows the evaluation results of the proposed OutFlip and other baseline systems. As can be observed from the table, the proposed OutFlip outperforms other baselines when the number of known intents is small. The small number of known intents is the most similar case to real-world applications, since in the open-world environment, the number of unknown intents is much larger than the number of known intents. The OutFlip also gives comparable results for ATIS and SNIPS corpus with the larger number of known intents.

For the Kakao corpus, the OutFlip performance is lower compared to the other baselines. To figure out the reason, the qualities of English GloVe embeddings and Korean GloVe embeddings are compared. We used 4 out of 14 categories in the word analogy corpus \citep{word2vec} for fair comparison; \texttt{capital-common-countries}, \texttt{capital-world}, \texttt{currency} and a subset of \texttt{family}. We removed all the syntactic questions since they cannot be translated into Korean words one-to-one. Part of \texttt{family} category is removed for the same reason. We also removed categories that give an advantage on English pre-trained embeddings; for example, the \texttt{city-in-state} category is removed because it contains relationships between US cities and US states. The remaining 6,168 questions are manually translated into Korean.

Table \ref{tbl:wordanalogy} shows the evaluation results of the English and Korean GloVe vectors on our subset of the word analogy corpus. As can be observed, the accuracy of Korean GloVe is much lower compared to the English GloVe vectors. Since the OutFlip relies on the cosine similarities between pre-trained embedding vectors to generate the OOD samples, the quality of embedding vectors is critical to the OOD generation performance.

Next, we applied the proposed OutFlip to other OOD detection baselines to see if the OutFlip could further improve their performance. Table \ref{tbl:result2} shows the evaluation results. In most cases applying the OutFlip to other OOD detection approaches resulted in performance improvement. The performance improvement was significant when the dataset is balanced, and the number of known intents is small.

We conducted a set of experiments to find out the best OutFlip iteration number and $T_{sim}$ value. Figure \ref{fig:test_sim} and Figure \ref{fig:test_attack} shows the OutFlip performances on the benchmark datasets with changing $T_{sim}$ values and OutFlip iteration numbers, respectively. Increasing $T_{sim}$ value would cause the OutFlip to generate more challenging examples, but the chance of developing wrong OOD samples would also increase.

Figure \ref{fig:test_sim} suggests that the balanced dataset like SNIPS could easily recover the errors introduced from large $T_{sim}$ values. In contrast, the ATIS dataset's macro F1 score decreases with the increased $T_{sim}$ values when many intents are known. Since 3 out of 18 ATIS intents have less than ten sentences, one or two erroneous OOD samples could lead to a performance drop. The macro F1 score of the balanced Kakao dataset does not increase with the $T_{sim}$ values larger than 0.3. Since the quality of Korean GloVe is relatively low, large $T_{sim}$ values introduce more errors compared to the English datasets.

As can be observed from Figure \ref{fig:test_attack}, in most cases, the macro F1 score converges with two to three OutFlip iterations. Additional OutFlip iterations give small or no performance improvements for balanced datasets and decrease macro F1 score for unbalanced dataset ATIS by introducing more errors.

\begin{table*}[t]
\centering
\begin{tabular}{ll|ccc|ccc} \hline
\multirow{2}{*}{Model}&Reference model &\multicolumn{3}{|c|}{\textbf{ATIS}} & \multicolumn{3}{c}{\textbf{SNIPS}}  \\ 
&for OutFlip&25\%&50\%&75\%&25\%&50\%&75\%\\ \hline
\multirow{3}{*}{BERT-base}&\textit{None}& 62.14 & 74.73 & 72.61 & 34.66 & 60.76 & 70.49 \\ 
&\textbf{OutFlip$_{cnn}$} & 70.60 & 76.08 & 72.45 & \textbf{74.51}& \textbf{81.09} & \textbf{86.92}  \\ 
&\textbf{OutFlip$_{lstm}$} & \textbf{71.58} & \textbf{78.22} &\textbf{72.68} & 73.02 & 81.04 &86.50 \\ \hline
\multirow{3}{*}{BERT-large}&\textit{None}& 62.17 & 74.01  & 73.32 &34.56 & 61.07 & 70.51  \\ 
&\textbf{OutFlip$_{cnn}$} & \textbf{76.02} & 80.68 &\textbf{79.54} & 70.95 & \textbf{80.53} & \textbf{87.06} \\ 
&\textbf{OutFlip$_{lstm}$} & 74.55 & \textbf{81.46} & 75.38&  \textbf{73.15}& 80.31 & 86.60\\ \hline

\end{tabular}
\caption{Results of applying OutFlip-generated samples to train models other than the reference model. Reference model \textit{None} means no OutFlip-generated OOD samples are added during training. }
\label{tbl:bertresult}
\end{table*}

One advantage of the OutFlip is that the generated OOD samples could be used to train and improve the OOD detection performance of the models other than the reference model without applying additional OutFlip iterations. We trained the BERT-base and BERT-large models  \citep{devlin2018bert} with the ATIS and SNIPS benchmarks. As the same as previous experiments, the unknown intents are removed during training and integrated back during testing. Besides, we added OOD samples generated using reference models \textbf{OutFlip$_{cnn}$} and \textbf{OutFlip$_{lstm}$} with three OutFlip iterations and $T_{sim}$ value 0.3, while training the BERT models.

Table \ref{tbl:bertresult} shows the evaluation results of the BERT models trained with the OutFlip-generated OOD samples. As can be observed from the table, the OutFlip-generated OOD samples significantly improved the OOD detection performances of BERT models, regardless of the reference models used to generate the OOD samples.

\subsection{Error Analysis}

\begin{table}
\centering
\begin{tabular}{lrrrr} \hline
\textbf{Iteration}  &\multicolumn{1}{c}{1} & \multicolumn{1}{c}{2} & \multicolumn{1}{c}{3} & \multicolumn{1}{c}{4}  \\ \hline
\textbf{ATIS} & 2003 & 1648.5 & 755.5 & 336.6 \\ 
\textbf{SNIPS} & 4923.5 &4644.2 & 3221.5 & 1956.2  \\ \hline
\end{tabular}
\caption{The number of generated OOD samples for each OutFlip iteration with reference model \textbf{OutFlip$_{cnn}$}, $T_{sim}=0.3$. 75\% of the intents are known. Numbers are the average of 10 known intent selections.  }
\label{tbl:geninstnum}
\end{table}

We randomly selected 200 samples from the OOD examples generated by \textbf{OutFlip$_{cnn}$} with three iterations and $T_{sim}=0.3$ for the ATIS dataset, when 75\% of the intents are known. The number of newly generated OOD samples for each iteration is shown in Table \ref{tbl:geninstnum}. We manually analyzed the selected examples for errors.

Among the 200 examples, 186 of them were correctly generated OOD sentences. Out of 14 error cases, 12 were due to the wrongfully extracted Core Class Tokens. Some ATIS intents have too few examples to extract Core Class Tokens; for example, intent \texttt{atis\_restriction} has only six samples. Also, an entity which shows up too frequently could also lead to wrong Core Class Token extraction result. For intent \texttt{atis\_flight}, 797 out of 4,334 samples contain the entity ``Denver".

For one case, the OutFlip-generated sentence accidentally belongs to the other in-domain intent. However, due to the reference model's error, the OutFlip fails to remove the generated sentence. A training instance ``Can you list the cheapest round trip \textbf{fare} from Orlando to Kansas City" (truth label \texttt{atis\_airfare} ) is converted to a sentence ``Can you list the cheapest round trip \textbf{airplane} from Orlando to Kansas City" (truth label \texttt{atis\_flight}), but the reference model classifies the converted sentence to \texttt{atis\_airfare}. Since the classification result remains the same, the OutFlip considers the generated sentence as ``hard-enough'' OOD sample.

The ATIS dataset allows an instance to have multiple labels; two or more labels are assigned to 23 ATIS training instances. The OutFlip failed to properly handle those instances. The remaining one error case is generated from a training instance with two assigned labels.

\section{Conclusion}

In this paper, we proposed OutFlip, a method to generate OOD samples using only in-domain training dataset. Our evaluation results showed that the proposed OutFlip could significantly improve the OOD detection performance of an intent classification model by iteratively generating difficult OOD samples. Since OutFlip does not require any modifications to model architecture, it could be used with other OOD detection approaches to improve OOD detection performance further. We also showed that the generated OOD samples could be used to train and improve the OOD detection performance of models other than the reference model, without applying additional OutFlip iterations.

Currently, we only focused on generating difficult OOD samples, which can fool the reference model. However, generating meaningful OOD samples could also be beneficial, since then the dialogue engine developer could check the generated OOD samples to find new intents. As our future work, we will focus on generating meaningful, fluent OOD samples.
\bibliography{acl21_ood}

\begin{thebibliography}{26}
\expandafter\ifx\csname natexlab\endcsname\relax\def\natexlab#1{#1}\fi

\bibitem[{Alzantot et~al.(2018)Alzantot, Sharma, Elgohary, Ho, Srivastava, and
  Chang}]{alzantot:18}
Moustafa Alzantot, Yash Sharma, Ahmed Elgohary, Bo-Jhang Ho, Mani Srivastava,
  and Kai-Wei Chang. 2018.
\newblock Generating natural language adversarial examples.
\newblock In \emph{Proceedings of the 2018 Conference on Empirical Methods in
  Natural Language Processing}, pages 2890--2896.

\bibitem[{Breunig et~al.(2000)Breunig, Kriegel, Ng, and Sander}]{lof}
Markus~M. Breunig, Hans-Peter Kriegel, Raymond~T. Ng, and Jörg Sander. 2000.
\newblock Lof: identifying density-based local outliers.
\newblock In \emph{Proceedings of the 2000 ACM SIGMOD international conference
  on Management of data.}, pages 93--104.

\bibitem[{Choi et~al.(2020)Choi, Park, Shin, Kim, and Shin}]{kakao}
DongHyun Choi, IlNam Park, Myeong~Cheol Shin, EungGyun Kim, and Dong~Ryeol
  Shin. 2020.
\newblock Integrated eojeol embedding for erroneous sentence classification in
  korean chatbots.
\newblock \emph{Computing Research Repository}, arXiv:2004.05744.

\bibitem[{Coucke et~al.(2018)Coucke, Saade, Ball, Bluche, Caulier, Leroy,
  Doumouro, Gisselbrecht, Caltagirone, Lavril, Primet, and Dureau}]{snips}
Alice Coucke, Alaa Saade, Adrien Ball, Théodore Bluche, Alexandre Caulier,
  David Leroy, Clément Doumouro, Thibault Gisselbrecht, Francesco Caltagirone,
  Thibaut Lavril, Maël Primet, and Joseph Dureau. 2018.
\newblock Snips voice platform: an embedded spoken language understanding
  system for private-by-design voice interfaces.
\newblock \emph{Computing Research Repository}, arXiv:1805.10190.

\bibitem[{Devlin et~al.(2018)Devlin, Chang, Lee, and
  Toutanova}]{devlin2018bert}
Jacob Devlin, Ming-Wei Chang, Kenton Lee, and Kristina Toutanova. 2018.
\newblock Bert: Pre-training of deep bidirectional transformers for language
  understanding.
\newblock \emph{arXiv preprint arXiv:1810.04805}.

\bibitem[{Ebrahimi et~al.(2018)Ebrahimi, Rao, Lowd, and Dou}]{hotflip}
Javid Ebrahimi, Anyi Rao, Daniel Lowd, and Dejing Dou. 2018.
\newblock Hotflip: White-box adversarial examples for text classification.
\newblock In \emph{Proceedings of the 56th Annual Meeting of the Association
  for Computational Linguistics (Volume 2: Short Papers)}, pages 31--36.

\bibitem[{Fei and Liu(2016)}]{Fei:16}
Geli Fei and Bing Liu. 2016.
\newblock Breaking the closed world assumption in text classification.
\newblock In \emph{Proceedings of the 2016 Conference of the North American
  Chapter of the Association for Computational Linguistics: Human Language
  Technologies}, pages 506--514.

\bibitem[{Goodfellow et~al.(2014)Goodfellow, Pouget-Abadie, Mirza, Xu,
  Warde-Farley, Ozair, Courville, and Bengio}]{gan}
Ian Goodfellow, Jean Pouget-Abadie, Mehdi Mirza, Bing Xu, David Warde-Farley,
  Sherjil Ozair, Aaron Courville, and Yoshua Bengio. 2014.
\newblock Generative adversarial nets.
\newblock \emph{Advances in neural information processing systems},
  27:2672--2680.

\bibitem[{Hemphill et~al.(1990)Hemphill, Godfrey, and Doddington}]{atis}
Charles~T. Hemphill, John~J. Godfrey, and George~R. Doddington. 1990.
\newblock The atis spoken language systems pilot corpus.
\newblock In \emph{Speech and Natural Language: Proceedings of a Workshop Held
  at Hidden Valley, Pennsylvania, USA, June 24-27}.

\bibitem[{Hendrycks and Gimpel(2017)}]{hendrycks:17}
Dan Hendrycks and Kevin Gimpel. 2017.
\newblock A baseline for detecting misclassified and out-of-distribution
  examples in neural networks.
\newblock In \emph{Proceedings of International Conference on Learning
  Representations}.

\bibitem[{Hochreiter and Schmidhuber(1997)}]{lstm}
Sepp Hochreiter and Jürgen Schmidhuber. 1997.
\newblock Long short-term memory.
\newblock \emph{Neural computation}, 9(8):1735--1780.

\bibitem[{Jin et~al.(2020)Jin, Jin, Zhou, and Szolovits}]{jin:20}
Di~Jin, Zhijing Jin, Joey~Tianyi Zhou, and Peter Szolovits. 2020.
\newblock Is bert really robust? a strong baseline for natural language attack
  on text classification and entailment.
\newblock In \emph{Proceedings of the AAAI conference on artificial
  intelligence}, volume~34, pages 8018--8025.

\bibitem[{Kim and Kim(2018)}]{kim:18}
Joo-Kyung Kim and Young-Bum Kim. 2018.
\newblock Joint learning of domain classification and out-of-domain detection
  with dynamic class weighting for satisficing false acceptance rates.
\newblock In \emph{Proceedings of Interspeech 2018}, pages 556--560.

\bibitem[{Kim(2014)}]{cnn-sen}
Yoon Kim. 2014.
\newblock Convolutional neural networks for sentence classification.
\newblock In \emph{Proceedings of the 2014 Conference on Empirical Methods in
  Natural Language Processing (EMNLP),}, pages 1292--1302.

\bibitem[{Kingma and Ba(2015)}]{adam}
Diederik~P. Kingma and Jimmy Ba. 2015.
\newblock Adam: A method for stochastic optimization.
\newblock In \emph{International Conference on Learning Representations},
  volume~5.

\bibitem[{Larson et~al.(2019)Larson, Mahendran, Peper, Clarke, Lee, Hill,
  Kummerfeld, Leach, Laurenzano, Tang, and Mars}]{larson:19}
Stefan Larson, Anish Mahendran, Joseph~J. Peper, Christopher Clarke, Andrew
  Lee, Parker Hill, Jonathan~K. Kummerfeld, Kevin Leach, Michael~A. Laurenzano,
  Lingjia Tang, and Jason Mars. 2019.
\newblock An evaluation dataset for intent classification and out-of-scope
  prediction.
\newblock In \emph{Proceedings of the 2019 Conference on Empirical Methods in
  Natural Language Processing and the 9th International Joint Conference on
  Natural Language Processing (EMNLP-IJCNLP)}, pages 1311--1316.

\bibitem[{Li et~al.(2020)Li, Ma, Guo, Xue, and Qiu}]{bert-attack}
Linyang Li, Ruotian Ma, Qipeng Guo, Xiangyang Xue, and Xipeng Qiu. 2020.
\newblock Bert-attack: Adversarial attack against bert using bert.
\newblock In \emph{Proceedings of the 2020 Conference on Empirical Methods in
  Natural Language Processing}, pages 6193--6202.

\bibitem[{Lin and Xu(2019)}]{LMCL-Lin:19}
Ting-En Lin and Hua Xu. 2019.
\newblock Deep unknown intent detection with margin loss.
\newblock In \emph{Proceedings of the 57th Annual Meeting of the Association
  for Computational Linguistics}, pages 5491--5496.

\bibitem[{Mikolov et~al.(2013)Mikolov, Chen, Corrado, and Dean}]{word2vec}
Tomas Mikolov, Kai Chen, Greg Corrado, and Jeffrey Dean. 2013.
\newblock Efficient estimation of word representations in vector space.
\newblock volume arXiv:1301.3781.

\bibitem[{Nalisnick et~al.(2018)Nalisnick, Matsukawa, Teh, Gorur, and
  Lakshminarayanan}]{lmcl-orig}
Eric Nalisnick, Akihiro Matsukawa, Yee~Whye Teh, Dilan Gorur, and Balaji
  Lakshminarayanan. 2018.
\newblock Do deep generative models know what they don't know?
\newblock In \emph{International Conference on Learning Representations.}

\bibitem[{Pennington et~al.(2014)Pennington, Socher, and Manning}]{glove}
Jeffrey Pennington, Richard Socher, and Christopher~D. Manning. 2014.
\newblock \href {http://www.aclweb.org/anthology/D14-1162} {Glove: Global
  vectors for word representation}.
\newblock In \emph{Empirical Methods in Natural Language Processing (EMNLP)},
  pages 1532--1543.

\bibitem[{Ryu et~al.(2018)Ryu, Koo, Yu, and Lee}]{ryu:18}
Seonghan Ryu, Sangjun Koo, Hwanjo Yu, and Gary~Geunbae Lee. 2018.
\newblock Out-of-domain detection based on generative adversarial network.
\newblock In \emph{Proceedings of the 2018 Conference on Empirical Methods in
  Natural Language Processing}, pages 714--718.

\bibitem[{Shu et~al.(2017)Shu, Xu, and Liu}]{doc-Shu:17}
Lei Shu, Hu~Xu, and Bing Liu. 2017.
\newblock Doc: Deep open classification of text documents.
\newblock In \emph{Proceedings of the 2017 Conference on Empirical Methods in
  Natural Language Processing}, pages 2911--2916.

\bibitem[{Xu et~al.(2020)Xu, He, Yan, Liu, Liu, and Xu}]{Xu:2020}
Hong Xu, Keqing He, Yuanmeng Yan, Sihong Liu, Zijun Liu, and Weiran Xu. 2020.
\newblock A deep generative distance-based classifier for out-of-domain
  detection with mahalanobis space.
\newblock In \emph{Proceedings of the 28th International Conference on
  Computational Linguistics}, pages 1452--1460.

\bibitem[{Yan et~al.(2020)Yan, Fan, Li, Liu, Zhang, Wu, and
  Lam}]{gaussian-Yan:20}
Guangfeng Yan, Lu~Fan, Qimai Li, Han Liu, Xiaotong Zhang, Xiao-Ming Wu, and
  Albert~Y.S. Lam. 2020.
\newblock Unknown intent detection using gaussian mixture model with an
  application to zero-shot intent classification.
\newblock In \emph{Proceedings of the 58th Annual Meeting of the Association
  for Computational Linguistics}, pages 1050--1060.

\bibitem[{Zheng et~al.(2020)Zheng, Chen, and Huang}]{zheng:20}
Yinhe Zheng, Guanyi Chen, and Minlie Huang. 2020.
\newblock Out-of-domain detection for natural language understanding in dialog
  systems.
\newblock \emph{IEEE/ACM Transactions on Audio, Speech, and Language
  Processing}, pages 1198--1209.

\end{thebibliography}
\bibliographystyle{acl_natbib}


\end{document}